\documentclass{article}

\usepackage{arxiv}

\usepackage[utf8]{inputenc} 
\usepackage[T1]{fontenc}    
\usepackage{hyperref}       
\usepackage{url}            
\usepackage{booktabs}       
\usepackage{amsfonts}       
\usepackage{nicefrac}       
\usepackage{microtype}      
\usepackage{lipsum}
\usepackage{amssymb,amsmath}
\usepackage{amsthm}
\usepackage{xspace}

\newcommand{\AF}{\ensuremath{\mathsf{AF}}\xspace}						
\newcommand{\arguments}{\ensuremath{\mathsf{A}}\xspace}				
\newcommand{\attacks}{\ensuremath{\mathsf{R}}\xspace}				
\newcommand{\AFcomplete}{\ensuremath{\AF=(\arguments,\attacks)}\xspace}		

\newcommand{\co}{\ensuremath{CO}}
\newcommand{\gr}{\ensuremath{GR}}
\newcommand{\st}{\ensuremath{ST}}
\newcommand{\pr}{\ensuremath{PR}}
\newcommand{\id}{\ensuremath{ID}}

\newcommand{\fudge}{\textsc{fudge}}

\newtheorem{theorem}{Theorem}

\title{fudge: a light-weight solver for abstract argumentation based on SAT reductions}

\author{
  Matthias Thimm\\
  Institute for Web Science and Technologies\\
  University of Koblenz-Landau,
  Germany\\
  \texttt{thimm@uni-koblenz.de} \\
  \And
  Federico Cerutti\\
  Department of Information Engineering\\
  University of Brescia,
  Italy\\
  \texttt{federico.cerutti@unibs.it} \\
  \And
  Mauro Vallati\\
  School of Computing and Engineering\\
  University of Huddersfield,
  United Kingdom\\
  \texttt{m.vallati@hud.ac.uk} \\
}

\begin{document}
\maketitle

\begin{abstract}
We present \fudge, an abstract argumentation solver that tightly integrates satisfiability solving technology to solve a series of abstract argumentation problems. While most of the encodings used by \fudge\ derive from standard translation approaches, \fudge\ makes use of completely novel encodings to solve the skeptical reasoning problem wrt.\ preferred semantics and problems wrt.\ ideal semantics.
\end{abstract}

\section{Introduction}
An \emph{abstract argumentation framework} $\AF$ is a tuple $\AF=(\arguments,\attacks)$ where \arguments is a (finite) set of arguments and \attacks is a relation $\attacks\subseteq \arguments\times\arguments$ \cite{Dung:1995}.
 For two arguments $a,b\in\arguments$ the relation $a \attacks b$ means that argument $a$ attacks argument $b$. For a set $S\subseteq \arguments$ we define
\begin{align*}
    S^+ & = \{a\in \arguments\mid \exists b\in S, b\attacks a\} \\
    S^- & = \{a\in \arguments\mid \exists b\in S, a\attacks b\}
\end{align*}
We say that a set $S\subseteq\arguments$ is \emph{conflict-free} if for all $a,b\in S$ it is not the case that $a\attacks b$.
A set $S$ \emph{defends} an argument $b\in\arguments$ if for all $a$ with $a\attacks b$ there is $c\in S$ with $c\attacks a$.
A conflict-free set $S$ is called \emph{admissible} if $S$ defends all $a\in S$. 

Different semantics \cite{Baroni:2011} can be phrased by imposing constraints on admissible sets. 
In particular, set $E$
\begin{itemize}
   	 \item is a \emph{complete} (\co) extension iff it is admissible and for all $a\in \arguments$, if $E$ defends $a$ then $a\in E$,
	\item is a \emph{grounded} (\gr) extension iff it is complete and minimal,
	\item is a \emph{stable} (\st) extension iff it is conflict-free and $E\cup E^+ =\arguments$,
	\item is a \emph{preferred} (\pr) extension iff it is admissible and maximal.
	\item is an \emph{ideal} (\id) extension iff $E\subseteq E'$ for each preferred extension $E'$ and $E$ is maximal.
\end{itemize}
All statements on minimality/maximality are meant to be with respect to set inclusion.

Given an abstract argumentation framework $\AFcomplete$ and a semantics $\sigma\in\{\co,\gr,\st,\pr,\id\}$ we are interested in the following computational problems\footnote{\url{http://argumentationcompetition.org/2021/SolverRequirements.pdf}}:
\begin{description}
    \item \textsf{SE-$\sigma$}: For a given abstract argumentation framework $\AF$, compute some $\sigma$-extension.
    \item \textsf{CE-$\sigma$}: For a given abstract argumentation framework $\AF$, determine the number of all $\sigma$-extensions.
	\item \textsf{DC-$\sigma$}: For a given abstract argumentation framework $\AF$ and an argument $a$, decide whether $a$ is in at least one $\sigma$-extension of $\AF$.
	\item \textsf{DS-$\sigma$}: For a given abstract argumentation framework $\AF$ and an argument $a$, decide whether $a$ is in all $\sigma$-extensions of $\AF$.
\end{description}
Note that \textsf{DC-$\sigma$} and \textsf{DS-$\sigma$} are equivalent for $\sigma\in\{\gr,\id\}$ as those extensions are uniquely defined \cite{Baroni:2011}. For these, we will only consider \textsf{DS-$\sigma$}.

The \fudge\ solver supports solving the above-mentioned computational problems wrt.\  all $\sigma\in\{\co,\gr,\st,\pr,\id\}$.
In the remainder of this system description, we give a brief overview on the architecture of \fudge\ (Section~\ref{sec:architecture}) and conclude in Section~\ref{sec:summary}.

\section{Architecture}\label{sec:architecture}
\fudge\ follows the standard reduction-based approach to solve the above-mentioned reasoning problems \cite{Besnard:2014,Cerutti:2018a} with the target formalism being the satisfiability problem \textsf{SAT} \cite{Biere:2009}. For example, given the problem \textsf{SE-ST} and an input argumentation framework $\AFcomplete$, first, for each argument $a\in \arguments$, we create a propositional variable $\text{in}_{a}$, with the meaning that $\text{in}_{a}$ is true in a satisfying assignment iff the argument $a$ is in the stable extension to be found. Then conflict-freeness can be modelled by the formula
\begin{align*}
	\Phi_1(\AF) & = \bigwedge_{(a,b)\in \attacks} \neg (\text{in}_{a}\wedge \text{in}_{b})
\end{align*}
while the constraint that all arguments not included in the extension must be attacked can be modelled by
\begin{align*}
    \Phi_2(\AF) & = \bigwedge_{a\in \arguments}(\neg\text{in}_{a}\Leftrightarrow\bigvee_{(b,a)\in R}\text{in}_{b})
\end{align*}
Then the formula $\Phi_1(\AF)\wedge \Phi_2(\AF)$ is satisfiable iff $\AF$ has a stable extension and a stable extension can be easily extracted from a satisfying assignment of $\Phi_1(\AF)\wedge \Phi_2(\AF)$. All reasoning problems on the first level of the polynomial hierarchy \cite{Dvorak:2018} can be solved in a similar manner and their corresponding counting problems can be realised by iterative satisfiability tests to enumerate all extensions.

Particularly challenging problems are those wrt.\ preferred semantics as, in particular, $DS$-$PR$ is $\Pi_2^P$-complete \cite{Dvorak:2018}. To solve that problem, we use the approach recently presented in \cite{Thimm:2021}. This approach relies on the following observation\footnote{Define $S'\attacks S$ iff there is $a\in S'$ and $b\in S$ with $(a,b)\in\attacks$.}:
\begin{theorem}[\cite{Thimm:2021}]
    $a\in\arguments$ is skeptically accepted wrt.\ preferred semantics iff
    \begin{enumerate}
        \item there is an admissible set $S$ with $a\in S$ and
        \item for every admissible set $S$ with $a \in S$ and every admissible set $S'$ with $S'\attacks S$, there is an admissible set $S''$ with $S' \cup \{a\} \subseteq S''$.
    \end{enumerate}
\end{theorem}
The above theorem states that we can decide skeptical acceptance wrt.\ preferred semantics by considering only those admissible sets that attack an admissible set containing the argument in question. As an admissibility check can be solved by a satisfiability check, similarly as above, the above insight leads to an algorithm that can solve $DS$-$PR$ without actually computing preferred extensions. The algorithm is presented in detail in \cite{Thimm:2021} and experiments confirm a significant performance improvement compared to previous encoding approaches.

Coming to ideal semantics, it is worth recalling \cite[Theorem 3.3]{ideal}.

\begin{theorem}[{\cite[Theorem 3.3]{ideal}}]
    An admissible set of arguments $S$ is ideal iff for each argument $a$ attacking $S$ there exists no admissible set of arguments containing $a$.
\end{theorem}

The interesting aspect of \cite[Theorem 3.3]{ideal} is that ideal semantics, although defined based on skeptical acceptance wrt.\ preferred semantics, does not rely on that notion. Moreover, in \cite{Thimm:2021} we also prove that starting from the set of arguments which are not attacked by an admissible set, its largest admissible set is the ideal extension. We can therefore tweak the machinery we created for computing skeptical acceptance wrt.\ preferred semantics to compute the ideal extension too. The complete algorithm is presented in \cite{Thimm:2021}.

\fudge\ is written in C++ and uses the satisfiability solver CaDiCaL 1.3.1\footnote{\url{http://fmv.jku.at/cadical/}} via its C++ API. CaDiCaL supports \emph{incremental} solving which provides a significant performance boost when successive satisfiability checks have to be made, in particular for counting problems. 

\section{Summary}\label{sec:summary}
We presented \fudge\, a reduction-based solver for various problems in abstract argumentation. \fudge\ leverages on a mix of standard and novel SAT encodings to solve reasoning problems, with the aim of avoiding the costly maximisation step that is characteristic of some of the abstract argumentation problems. 
The source code of \fudge\ is available at \url{http://taas.tweetyproject.org}.

\bibliographystyle{unsrt}  
\bibliography{references.bib}

\begin{thebibliography}{1}

\bibitem{Dung:1995}
Phan~Minh Dung.
\newblock {O}n the {A}cceptability of {A}rguments and its {F}undamental {R}ole
  in {N}onmonotonic {R}easoning, {L}ogic {P}rogramming and n-{P}erson {G}ames.
\newblock {\em Artificial Intelligence}, 77(2):321--358, 1995.

\bibitem{Baroni:2011}
Pietro Baroni, Martin Caminada, and Massimiliano Giacomin.
\newblock An introduction to argumentation semantics.
\newblock {\em The Knowledge Engineering Review}, 26(4):365--410, 2011.

\bibitem{Besnard:2014}
Philippe Besnard, Sylvie Doutre, and Andreas Herzig.
\newblock Encoding argument graphs in logic.
\newblock In {\em International Conference on Information Processing and
  Management of Uncertainty in Knowledge-based Systems - IPMU 2014}, 2014.

\bibitem{Cerutti:2018a}
Federico Cerutti, Sarah~A. Gaggl, Matthias Thimm, and Johannes~P. Wallner.
\newblock Foundations of implementations for formal argumentation.
\newblock In Pietro Baroni, Dov Gabbay, Massimiliano Giacomin, and Leendert
  van~der Torre, editors, {\em Handbook of Formal Argumentation}, chapter~15.
  College Publications, February 2018.

\bibitem{Biere:2009}
Armin Biere, Marijn Heule, Hans van Maaren, and Toby Walsh, editors.
\newblock {\em Handbook of Satisfiability}, volume 185 of {\em Frontiers in
  Artificial Intelligence and Applications}.
\newblock IOS Press, 2009.

\bibitem{Dvorak:2018}
Wolfgang Dvo\v{r}\'{a}k and Paul~E. Dunne.
\newblock Computational problems in formal argumentation and their complexity.
\newblock In Pietro Baroni, Dov Gabbay, Massimiliano Giacomin, and Leendert
  van~der Torre, editors, {\em Handbook of Formal Argumentation}, chapter~14.
  College Publications, February 2018.

\bibitem{Thimm:2021}
Matthias Thimm, Federico Cerutti, and Mauro Vallati.
\newblock Skeptical reasoning with preferred semantics in abstract
  argumentation without computing preferred extensions.
\newblock In {\em Proceedings of the 30th International Joint Conference on
  Artificial Intelligence (IJCAI'21)}, 2021.

\bibitem{ideal}
P.~M. Dung, P.~Mancarella, and F.~Toni.
\newblock Computing ideal sceptical argumentation.
\newblock {\em Artificial Intelligence}, 171(10):642--674, 2007.

\end{thebibliography}
\end{document}